\documentclass[11pt,letterpaper]{article}

\usepackage[preprint]{acl}

 \usepackage{microtype}

\usepackage[english,bidi=default]{babel} 
\babelfont{rm}{TeXGyreTermesX} 

\usepackage{graphicx}
\usepackage{booktabs}
\usepackage{subcaption}
\usepackage{siunitx} 
\usepackage[shortcuts]{extdash} 

\title{AI as a Tool for Simulation-Based Experiments in Literary Studies}

\author{Matthew Wilkens \\
  Department of Information Science \\
  Cornell University \\
  \texttt{wilkens@cornell.edu}
}

\begin{document}
\maketitle
\begin{abstract}
 Generative artificial intelligence (AI) systems open new possibilities for experimentation in literary studies via controlled, grounded, large-scale, low-cost simulations of cultural production. Current systems have not yet been shown to produce high-quality, book-length narrative texts that reliably reflect arbitrarily specified cultural constraints or stylistic features. But there exists substantial relevant research on each of the components required for literary-historical simulation. These include the use and validation of AI systems as proxies for differentiable human populations; the narrative and stylistic properties of AI-generated texts; the stability and coherence of multiagent, multiturn AI simulations of human actors; and technical methods through which to alter in predictable ways the knowledge and behavior of generative systems. Together, these areas could provide a starting point for more ambitious AI-based modeling of cultural systems of literary production.

We describe the possibilities and challenges of simulation-based experiments in literary studies, summarize the current state of the art in relevant fields, and explain key technical aspects of the work. To provide an example directly relevant to literary scholars, we present the results of experiments on literary text generation, including comparisons to high-status, human-authored novels. Our results include the first demonstration of (limited) in-distribution outputs by AI models in this domain. We conclude with a description of future work on full counterfactual literary-historical simulations using AI.
\end{abstract}

\section{Introduction}

Neither history nor literary studies is defined by its reliance on laboratory-style experimentation. The reasons for this are obvious. It is hard to run experiments on people who are dead or on events that have already happened. Even in cases where one might design forward-looking studies (of contemporary authorship under varying exposure to economic precarity, for example), the costs and ethical issues involved are generally prohibitive.

The lack of viable experimental research in literary studies is unfortunate, because the ability to study in isolation the effects of specific interventions or counterfactual conditions via laboratory-style experiments, randomized controlled trials, or even survey methods, would provide new leverage on important problems. But literature and history are not alone in this difficulty. Many fields have learned to study phenomena they cannot directly manipulate. Climate scientists build computational models of the atmosphere to test the effects of greenhouse gas concentrations that have not (yet) occurred. Epidemiologists simulate disease transmission to evaluate interventions they cannot impose on real patients or communities. Economists populate agent-based models with virtual actors to examine market dynamics under conditions they cannot otherwise control. In each case, the logic is the same. When direct experimentation is impossible, impractical, or unethical, simulation provides a principled alternative. The simulation need not be a perfect replica of the target system; it must be good enough, along specified dimensions, to support useful inference.

Literary studies has not had access to the experimental toolkits enabled by simulation. We cannot rerun the twentieth century without James Joyce to see how modernism might have developed differently. We cannot expose matched cohorts of novelists to varying levels of economic success or deprivation to measure the effect on their formal choices. We cannot hold constant every feature of the postcolonial literary field except the migration patterns of Caribbean writers to isolate the influence of displacement on narrative form. These are counterfactual questions -- precisely the kind that experimentation is designed to address -- and literary scholars ask them constantly, in the form of claims about influence, context, and causation. They have also been explored as such in theoretical terms by scholars including Catherine Gallagher, Paul Saint-Amour, and Andrew Miller. But we have had no rigorous means of testing such claims.

This paper argues that generative AI systems are beginning to change this situation. Specifically, we propose that large language models (LLMs) can serve as instruments for simulation-based experiments in literary studies. The core idea is straightforward. If an LLM can be conditioned to produce texts that vary in measurable and predictable ways along dimensions relevant to literary analysis -- genre, period, cultural context, stylistic register -- then it should be possible to design experiments in which those dimensions are systematically manipulated while others are held constant. The resulting texts are not literature; they are simulated literary outputs grounded in the range of documentary sources on which LLMs are trained, useful for the same reasons that simulated weather is useful to meteorologists.

The present contribution is primarily methodological. We describe a framework for simulation-based literary experiments, situate it with respect to existing work in natural language processing (NLP), computational social science, and literary studies, and present the results of a large-scale empirical study comparing AI-generated fiction to a corpus of human-authored novels. The results are promising. AI-generated texts preserve genre structure and respond to prompt engineering in predictable directions. They are \emph{less} internally homogeneous than human fiction and show aggregate lexical signatures that distinguish them only weakly from human texts. Full-blown simulations, especially of contemporary literary-social processes that encounter fewer of the challenges inherent to historical modeling, may be on the edge of tenability.

The paper proceeds as follows. Section~\ref{sec:background} reviews related work across five domains: literary-theoretical analysis of counterfactual history, the use of LLMs as simulated human populations, the properties of AI-generated literature, model editing and unlearning for counterfactual conditioning, and validity frameworks for simulation research. Section~\ref{sec:design} describes the experimental design, including goals, corpus construction, synthetic author biographies, chapter generation pipelines, and document-embedding-based analysis methods. Section~\ref{sec:results} presents our empirical findings. Section~\ref{sec:discussion} interprets the results, addresses major challenges and limitations, and outlines a typology of research scenarios for simulation-based literary studies. We conclude in section~\ref{sec:conclusion} with a summary of contributions and an invitation to both literary and computational researchers.

\section{Background and related work}
\label{sec:background}

The proposal to use AI for simulation-based literary experiments draws on five overlapping areas of research. We review each briefly, emphasizing what has been validated, what remains uncertain, and what each contributes to the present project. 

If all you care about are the experiments we ran, you can skip to section~\ref{sec:design}. If you just want the results, they're in section~\ref{sec:results}.

\subsection{Theories of counterfactual history}

There was, in the late 2000s and early 2010s, a raft of both creative production and critical interest in counterfactual history (also called alternate history or allohistory), much of it related to the specific examples of the US Civil War and World War II. Catherine Gallagher's \emph{Telling It Like It Wasn't} \citeyearpar{gallagherTellingItIt2018} is the culminating instance of the critical genre, tracing the evolution of alternate history from a theological matter concerning divine providence to more prosaic questions in military and economic history. Gallagher's work is especially tied to a special forum published a decade earlier in \emph{Representations} that included work by \citeauthor{maslanTellingLiveTale2007, jainLivingPrognosisElegiac2007, millerLivesUnledRealist2007, saint-amourChristmasComeHospitality2007} and  \citeauthor{gallagherWhenDidConfederate2007} herself \citep[see also][]{saint-amourCounterfactualStatesAmerica2011a}.

In most of these cases, and certainly in Gallagher, counterfactual history represents a historical method devised to illuminate deductively the causal forces and mechanisms that govern a concrete historical nexus. So, for example, the reputable version of alternate history that explores a Confederate victory in the Civil War is dedicated to explaining the diplomatic, political, and economic factors that shaped the decades following the end of that war. It draws on period documents and the historian's sense of both the individual psychology of important actors and the network of forces to which they and the relevant institutions were (or would have been) subject. It is, in this regard, a variety of analysis that is continuous with most other academic history.

Simulation-based historical analysis of the type explored in this paper is epistemically continuous with counterfactual history in the mold of Gallagher or of mainstream historical explanation. It, too, explores why past situations developed as they did, in part by asking what would have been the case, and why, if some salient factor had been different. This is what it means to reason causally. The difference, in the case of simulation, is that such reasoning has access to a new type of evidence, namely the corpus-grounded probabilistic distribution of potential outcomes under unseen conditions. That distribution of outcomes can be, moreover, observed at reasonable cost, which in turn makes counterfactual historical reasoning attractive for cases that are either (or both) more complex or less consequential than the inverted outcomes of major wars. 

This argument concerning the epistemic compatibility of simulation with historical analysis is a kind of evolution of older debates about the value of computationally derived evidence in literary studies. Just as observational data from large corpora can be a useful element of critical arguments about aspects of literary history, including by opening new questions about which it was previously difficult or impossible to produce satisfactory evidence, so too can simulations of counterfactual conditions provide practically accessible evidentiary support for causal inferences about history. LLM-based simulations provide access to the range of knowledge, reasoning styles, and implied preferences present in their training data. For that reason, they capture implicit and distributed aspects of cultural situations about which it is otherwise difficult to make evidence-based, probabilistic arguments. To say the obvious, changing the types of evidence available, and the cost of producing them, necessarily changes the questions that are and are not worth pursuing.

\subsection{Silicon samples and simulated populations}

The most directly relevant precedent for AI-based simulation studies of cultural processes comes from the social sciences, where researchers have begun using LLMs as proxies for human survey respondents and experimental subjects. \citet{argyleOutOneMany2023} introduced the concept of ``silicon samples,'' demonstrating that GPT-3, when conditioned on sociodemographic backstories drawn from real survey participants, exhibits ``algorithmic fidelity,'' the ability to reproduce fine-grained, demographically correlated response distributions across a wide variety of human subgroups. \citet{hortonLargeLanguageModels2023} extended this logic to economics, showing that LLMs prompted toward specific preferences and endowments behave as plausible economic agents in replications of classic experiments, and proposing that such ``homo silicus'' agents could serve as pilot instruments for social science research.

Subsequent work has both strengthened and complicated these findings. \citet{kozlowskiSilicoSociologyForecasting2024} demonstrated that an LLM trained on texts published through October, 2019 (GPT-3) could reproduce observed partisan differences in COVID-19 policy preferences in 84\% of cases, suggesting that LLMs model ideological landscapes well enough to forecast opinion polarization. \citet{grossmannAITransformationSocial2023a} argued in \textit{Science} that AI can transform social science research, but emphasized that careful bias management and data fidelity are essential. \citet{dillionCanAILanguage2023a} provided a cautionary assessment of whether LLMs can truly replace human participants, noting important gaps in individual-level accuracy. Most recently, \citet{anthisLLMSocialSimulations2025} reviewed empirical comparisons between LLMs and human subjects, concluding that LLM social simulations are a promising research method and recommending context-rich prompting and fine-tuning with domain-specific datasets.

Two findings from this literature are especially relevant to literary simulation. First, conditioning matters. The same model produces very different outputs depending on the demographic, cultural, and contextual information provided in the prompt \citep{argyleOutOneMany2023, kozlowskiSilicoSociologyForecasting2024}. Second, aggregate distributional properties are more reliably reproduced than individual-level responses \citep{dillionCanAILanguage2023a, anthisLLMSocialSimulations2025}. Both points have direct implications for the design of literary experiments, where the goal is to study how populations of texts vary across conditions rather than to produce any single text that matches a specific human author.

No existing work in this tradition, however, has applied the silicon-samples framework to literary or creative production. The extension from survey-style opinion reproduction to the generation of literary texts under culturally specified conditions is a novel contribution of the present work.

\subsection{AI-generated literature}

A growing literature examines the narrative and stylistic properties of LLM-generated text. \citet{walshDoesChatGPTHave2024} compared 5,760 GPT-generated poems to 3,874 human poems, finding that, while GPT can produce formally correct verse in various structures, it exhibits strong stylistic uniformity, including a preference for rhyme, quatrains, iambic meter, first-person plural perspectives, and a narrow vocabulary. \citet{chakrabartyArtArtificeLarge2024} used a creativity assessment framework evaluated by professional writers and found that LLM-generated stories pass three to ten times fewer creativity tests than stories by successful professional authors. \citet{xuEchoesAIQuantifying2025}, in a study published in \emph{PNAS}, introduced a metric for measuring plot uniqueness and demonstrated that LLM-generated stories contain ``echoed'' plot elements across generations, while human-written plots are rarely recreated. \citet{zhangNoveltyBenchEvaluatingLanguage2025} reported that current models generate significantly less diversity than human writers, with larger models sometimes exhibiting less diversity than smaller ones. \citet{houCreativityPrismHolisticBenchmark2025} decomposed creativity into quality, novelty, and diversity, finding that strong performance on one dimension does not guarantee strong performance on others. \citet{russellStoryScopeInvestigatingIdiosyncrasies2026}, operationalizing the framework of \citet{hamilton-etal-2026-narrabench} with an eye toward AI detection, find that aspects of narrative structure are distinctively distributed in AI-generated stories compared to human-authored ones.

In short, prior work on AI-generated literature finds such literature to be both shallow and narrow. These findings establish important baselines, but they often frame the question differently than we do here. Where existing literature evaluates AI-generated text as a creative product, asking ``is this good literature?'' the simulation framework asks ``does this output vary in measurable and predictable ways that parallel how actual literary texts differ across specified conditions?'' A text need not be good fiction to be a useful simulation of fiction, just as a climate model need not produce pleasant weather to be informative about atmospheric dynamics. What matters is whether the simulated outputs preserve enough structural and distributional properties of the target domain to support valid inference about the effects of controlled interventions.

\subsection{Model training and tuning}

The most ambitious application of literary simulation -- counterfactual experiments that ask how literature would have differed under altered conditions -- requires the ability to modify what a model knows. Two technical problems are relevant in this case.

\textbf{Model editing} involves directly altering the factual associations stored in a model's weights. \citet{mengLocatingEditingFactual2023} developed ROME (Rank-One Model Editing), demonstrating that specific factual associations can be located in mid-layer feed-forward modules of deep neural networks and directly edited. \citet{mengMassEditingMemoryTransformer2023} extended this to MEMIT, which can simultaneously update thousands of associations. These techniques show that targeted knowledge manipulation is feasible in principle, though current methods are limited primarily to discrete factual associations rather than the distributed stylistic and cultural knowledge that would need to be altered for culture- or cohort-level literary and historical  counterfactual experiments.

\textbf{Model unlearning} attempts to remove specific knowledge from a trained model. \citet{eldanWhosHarryPotter2023} demonstrated that Harry Potter--related content could be effectively erased from a 7-billion-parameter model in approximately one GPU hour of fine-tuning, providing the closest proof-of-concept for literary knowledge removal. However, \citet{mainiTOFUTaskFictitious2024} developed a benchmark for evaluating unlearning and found that no existing method achieves truly effective unlearning, because the modified models do not behave as if the target data was never learned. This is a significant limitation for counterfactual simulation, which requires not merely suppressing references to a target but genuinely altering the model's generative behavior as if that knowledge were absent.

\textbf{Pretraining on historical sources} is an alternate or complementary approach that generally trades off model capability (due to limited training and post-training data) and cost against presumptively superior isolation of period knowledge and reasoning. While some such historical models have been targeted primarily toward pedagogical and public-engagement applications (``talk to Abraham Lincoln!''), serious efforts at research-relevant models have been reported by \citet{dgoettlichHistoryllmsRanke4b} and \citet{underwoodCanLanguageModels2025}. Very recently, \citet{IntroducingTalkie13B} reported the development of the largest historical language model to date, a 13B-parameter LLM trained on 260B tokens of English-language text published before 1931. While none of these models has yet been evaluated in full, they represent an additional route by which historically representative models might be constructed.

Separately, the challenge of historical fidelity constrains what temporal counterfactuals are currently possible. \citet{underwoodCanLanguageModels2025} found that prompting a contemporary model with period prose does not produce period-consistent output, and that fine-tuning produces results convincing enough to fool simple automated judges, but not expert human evaluators. They tentatively conclude that pretraining on period-specific text may be required. \citet{fittschenPretrainingLanguageModels2025} reached a similar conclusion via a different route, arguing that domain-restricted pretraining is the only reliable guarantee for restricting a model's outputs to a specific historical period (but note that \citeauthor{IntroducingTalkie13B} illustrates the practical difficulty of achieving historical isolation). \citet{atariHistoricalPsychology2023a} provided a broader framework for computational analysis of historical psychology, demonstrating the potential and the challenges of extracting period-specific information from large corpora.

Finally, we note the direct connection between simulation and causal inference. The counterfactual questions that motivate literary simulation -- how would the literary field have differed if some condition had been otherwise?\ -- are fundamentally causal questions. \citet{federCausalInferenceNatural2022a} provide a comprehensive survey of causal inference in NLP, covering settings where text functions as outcome, treatment, or confounding variable. \citet{keithTextCausalInference2020} review methods for using text to address confounding in causal estimates. These frameworks provide the formal language needed to move from abstract counterfactual speculation to proper experimental design.

\subsection{Validation and benchmarking}

Any simulation-based research program must address the question of validity, specifying and quantifying the conditions under which a simulation's output supports valid conclusions about the target system. Several recent contributions provide frameworks for thinking about this question.

\citet{Underwood_Qiu_Griebel_Nelson_Roland_Shang_Wilkens_2026} introduce a document-grounded benchmark of historical reasoning in LLMs, drawing both questions and answers from period sources and evaluating responses using a mix of multiple-choice accuracy, generative perplexity metrics, and LLM-as-judge methods. As the authors note, historical LLM evaluation poses special challenges compared to the contemporary case, since the evaluation task is, today, strictly superhuman (there are no living native informants concerning the texture of social experience before about 1930). For this reason, while modern expert evaluation of the historical fidelity of LLM outputs is desirable in many cases, it is not a sufficient gauge of model performance, even setting aside the cost of scaling it beyond specialized examples. A central role for comparison to elements of the surviving historical record is thus impossible to avoid, even acknowledging the gaps and biases of that record. It is here, in selecting benchmark sources that capture the widest possible range of relevant perspectives and experiences, that the contribution of experts is likely to be most critical, so that the resulting benchmarks assess those aspects of social situatedness that models might be used to simulate.\footnote{We note in passing that the interpretive study of LLM-generated literature in lieu of human-authored primary texts is a potentially fruitful, but almost entirely distinct, project from the evaluation of historical fidelity.}

In addition to model evaluation, \citet{zhouPIMMURPrinciplesEnsuring2025} articulate six methodological requirements for valid LLM-based social simulation, formalized as the PIMMUR principles: Profile (agents must be heterogeneous), Interaction (natural rather than scripted), Memory (retained across turns), Minimal-control (prompts should not predetermine outcomes), Unawareness (agents should not be able to infer the experimental hypothesis), and Realism (validation against real-world data rather than simplified theory). Strikingly, they find that GPT-4o and Qwen3 infer the underlying experiment in 53\% of cases they examined and that, when PIMMUR principles are enforced, many previously reported social phenomena fail to emerge. Care must be taken to design simulation experiments that are robust to these issues.

\citet{wallachPositionEvaluatingGenerative2025} argue that evaluating generative AI is fundamentally a social science measurement challenge and present a four-level framework grounded in measurement theory that brings rigor to debates about validity. \citet{almaatouqPlaying20Questions2024} propose an integrative experiment design framework that explicitly maps design spaces and iteratively generates and tests theories. Their approach maps naturally onto simulation-based literary research, where the design space of possible experiments is vast and largely unexplored.

Directly relevant to the present case is \citet{rolandSocialAIAuthor2025}, which simulates 101 ``AI authors'' modeled on real-world counterparts and analyzes how LLMs articulate literary distinction across race, gender, and publishing context. Their findings are important and concerning. Outputs are dominated by a reductive binary, intra-group diversity is flattened, and prompt-based interventions reduce racial fixation but often efface race altogether. \citet{lee2026pooralignmentsteerabilitylarge} reach similar conclusions with respect to simulated college application essays, in which they observe frequent collapse to stereotype when prompting AI models toward minoritarian identities. These studies provide the closest existing precedents for the present paper; their findings suggest that there exist significant barriers to human-like creative writing in the aggregate, even absent the challenges of historical fidelity.

\section{Experimental design}
\label{sec:design}

In response to the difficulties noted in prior work, we designed a large-scale experiment to test the viability of AI-generated literature as a distributional proxy for human-authored literature belonging to several genres. This experiment entails three primary elements: a corpus of human-authored novels and narrative nonfiction; a metric by which to evaluate the similarity and distributional properties of both the human- and AI-authored texts; and a prompt-based pipeline for text generation that builds on best practices from the existing literature. This section describes the details of the experimental setup.

\subsection{Corpus and metadata}

The human reference corpus is drawn from CONLIT, a collection of contemporary long-form narratives in English \citep{piperCONLITDatasetContemporary2022}. CONLIT contains 2,754 volumes, of which 1,934 represent novel-like fiction spanning nine genres of varying reception status (from prize-nominated literary novels to bestsellers, genre fiction, and young adult novels). The remaining 820 volumes are narrative nonfiction (histories, biographies, and memoirs). The full corpus spans the years 2001--2021, with most volumes concentrated between 2005 and 2018. The primary comparison set used in the experiment comprises 258 novels that were short-listed for, or winners of, major literary prizes including the Booker Prize, National Book Award, PEN/Faulkner Award, Giller Prize, and Governor General's Literary Award. These are identified by the label `PW' in the results below. Additional direct comparisons focus on the science fiction (SF) and mystery (MY) subsets of the corpus. There are 1,676 non-prize fiction volumes that are used as background data for lexical analysis. The corpus includes metadata on publication date, author demographics, genre classification, reception data, and other facets described at length in \citet{piperCONLITDatasetContemporary2022}.

\subsection{Chapter generation}

AI-generated texts were produced using GPT-5 (\texttt{gpt-5-2025-08-07}; \citealt{singh2025openaigpt5card}) via the OpenAI batch API under two conditions: complex prompts and basic prompts.

\subsubsection{Complex prompts} 

Complex prompts used a multi-component structure. A system prompt instructed the model to behave as an expert literary novelist, crafting stories that are emotionally and intellectually engaging, formally innovative, and thematically original. A user prompt asked the model to write the first chapter (approximately 5,000 words) of a new novel that might be competitive for a literary prize, preceded by an internal plan and character description following \citet{gurungLearningReasonLongForm2025} and \citet{gurung-lapata-2024-chiron} that would not appear in the output. Chapter-length generations were preferred to full-length version for reasons of cost and output coherence. While the quality of generations in excess of 30k tokens is rapidly improving, we sought to avoid the difficulties introduced by relying on output lengths known to suffer from degraded coherence \citep{que2024hellobenchevaluatinglongtext, hamilton2025longdidntmodeldecomposing}. We consider the reliability of this synechdochal substitution in section \ref{sec:embed}. 

Each generation received a randomly selected synthetic author biography (see the next section, below) as personal context, derived from the set of real authors in each target genre. Additionally, a subset received a historical date directive (``Respond as if you are writing in the year \ldots'') for years between 2005 and 2016, matching the distribution of dates in the PW set. Genre-specific variants of these prompts were created for science fiction and mystery. This pipeline produced 1,400 prize-style chapters (1,200 dated across twelve years, as well as 200 undated outputs not prompted for any specific target year), 450 science fiction chapters, and 450 mystery chapters, for a total of 2,300 complex-prompted generations, all using GPT-5. Examples of the prompts are included in Appendix~\ref{sec:prompts}. Full source code, generated texts, and analysis data are available at \url{https://github.com/wilkens/literary-generation}.

\subsubsection{Synthetic author biographies}
A distinctive feature of the experimental design is the use of synthetic author biographies to provide individuated cultural context for text generation. For each real author in the prize-nominated, science fiction, and mystery subsets of the corpus, a biographical sketch was generated using GPT-5. The prompt instructed the model to write an approximately 100-word biography of an imaginary author who closely resembles the named real author, written in the second person (``You were born in \ldots'') as if the current year were 2010. The biography was to include place and year of birth, a summary of life and career, distinguishing characteristics of the author's work, and at most one or two major awards. The details were to be similar but not identical to the real author's, without the use of fictitious places, awards, or book titles, and without inventing a name for the imaginary author.

The resulting biographies provide something that generic prompts lack: a specific, if lightly sketched, authorial subject position grounded in particular life experiences, cultural contexts, and literary trajectories. The synthetic biography for an author resembling Ian McEwan, for example, describes a writer born in 1950 in Portsmouth, raised abroad in a military family, educated at Bristol and Lancaster, who emerged in the mid-1970s with short stories noted for psychological intensity. This is not Ian McEwan, but it is an authorial persona with a similar, concrete social location and literary style, the type of individuating context that, according to the silicon-samples literature, improves the fidelity of LLM-based simulation.

\subsubsection{Basic prompts} 

Basic prompts used a minimal instruction, simply ``Write the first chapter of a \ldots\  novel,'' with no system prompt, no biographies, no dates, and no instructions about quality or planning. Basic prompts were tested across five models (GPT-5, GPT-5-mini, GPT-5-nano, GPT-4.1, and GPT-3.5-turbo), with 100 texts per genre per model, yielding 1,500 basic-prompted generations.

The total corpus of AI-generated texts comprises 3,800 chapters: 2,300 complex-prompted and 1,500 basic-prompted. Total API usage cost for all experiments was approximately 200 USD.

\subsection{Embedding and analysis\label{sec:embed}}

We assess text similarity using the standard metric of cosine similarity in document embedding space. All texts were embedded using Qwen3-Embedding-8B \citep{zhang2025qwen3embeddingadvancingtext}, a prompt-steerable 8-billion-parameter document embedding model with a 32k-token input window that produces 4,096-dimensional outputs. Embeddings were generated with a prompt emphasizing content, form, and style prepended to each chunk, designed to bias the embedding space toward features relevant to literary analysis rather than generic semantic similarity.

Human texts were divided into non-overlapping chunks of up to 32,000 tokens, with additional 2,000-token and 5,000-token start chunks used for validation. Book-level embeddings were computed as length-weighted means across all chunks, then $L^2$-normalized. AI-generated chapters, at an average of about 10,500 words (or 13,000 tokens) each, all fell within a single chunk.\footnote{Careful readers will note that the model was prompted to produce chapters of 5,000 words each. This instruction was not well followed in aggregate. Chapter lengths produced by GPT-5 varied approximately normally between 6,000 and 15,000 words. Chapters produced by earlier models and using basic prompts were generally shorter and were similarly distributed.}

To validate that chapter-length initial portions are representative of full-book embeddings, we computed similarities between document-initial chunks and full-book embeddings for human texts. The similarity between the first 5,000 or 32,000 tokens and the complete book is substantially higher than within-genre mean similarities, supporting the use of chapter-length AI texts as proxies for comparison.

For visualization only, the embedding space was reduced to two dimensions via UMAP \citep{mcinnes2020umapuniformmanifoldapproximation} fitted to human-authored texts; AI-generated texts were projected into the same space. This ensures that the coordinate system reflects the structure of human literary fiction, with AI texts mapped as external points. 

For lexical analysis, we used the Fightin' Words method of \citet{monroe2008fightin}, which identifies words statistically overrepresented in one corpus relative to another, using an informative Bayesian prior drawn from 1,676 non-prize-winning fiction volumes.

\section{Results}
\label{sec:results}

\subsection{Genre structure in embedding space}

\begin{figure*}[t]
    \centering
    \begin{subfigure}[t]{0.48\textwidth}
        \centering
        \includegraphics[width=\textwidth]{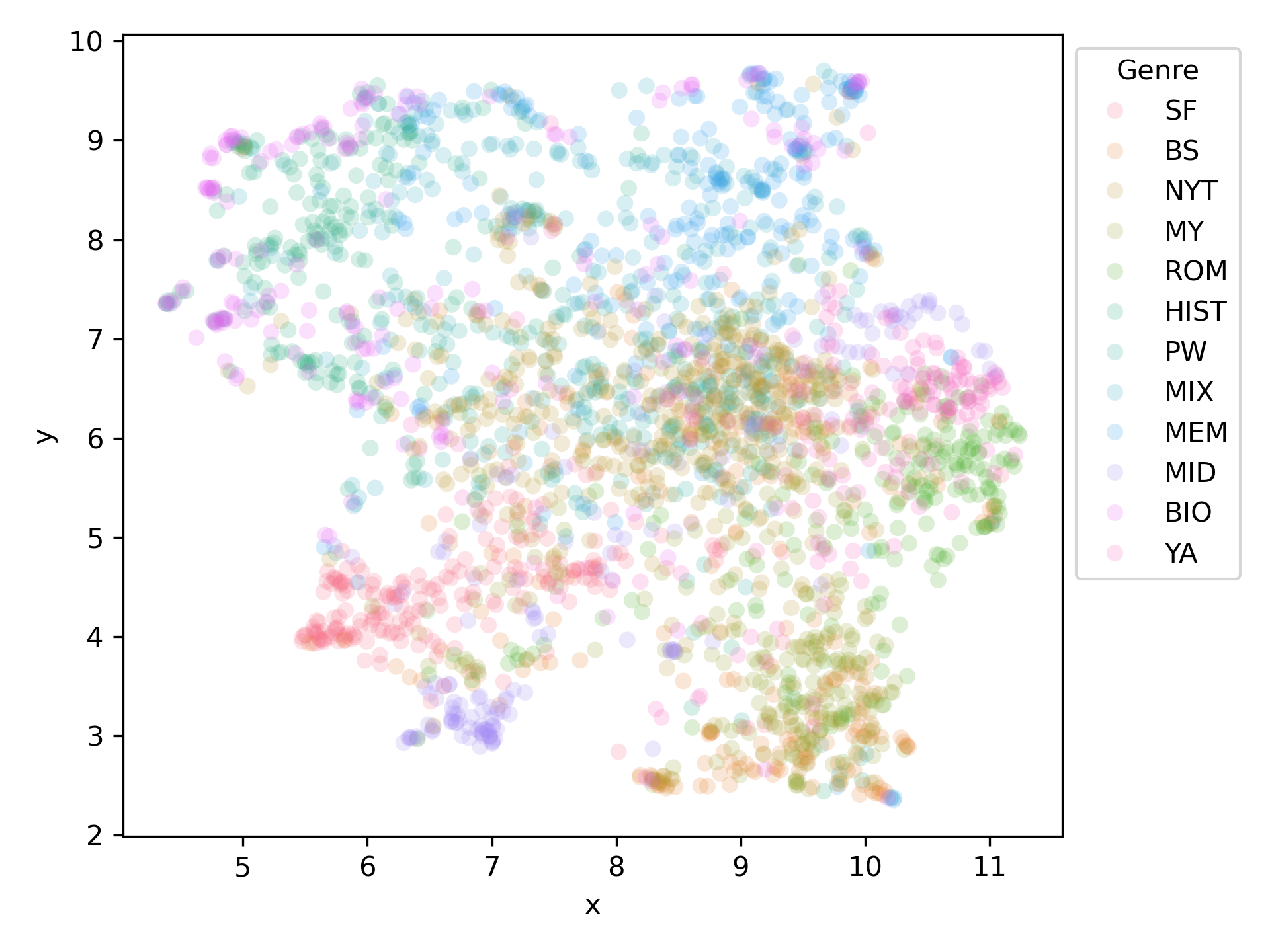}
        \caption{Human-authored narratives by genre}
        \label{fig:first_sub}
    \end{subfigure}
    \hfill
    \begin{subfigure}[t]{0.48\textwidth}
        \centering
        \includegraphics[width=\textwidth]{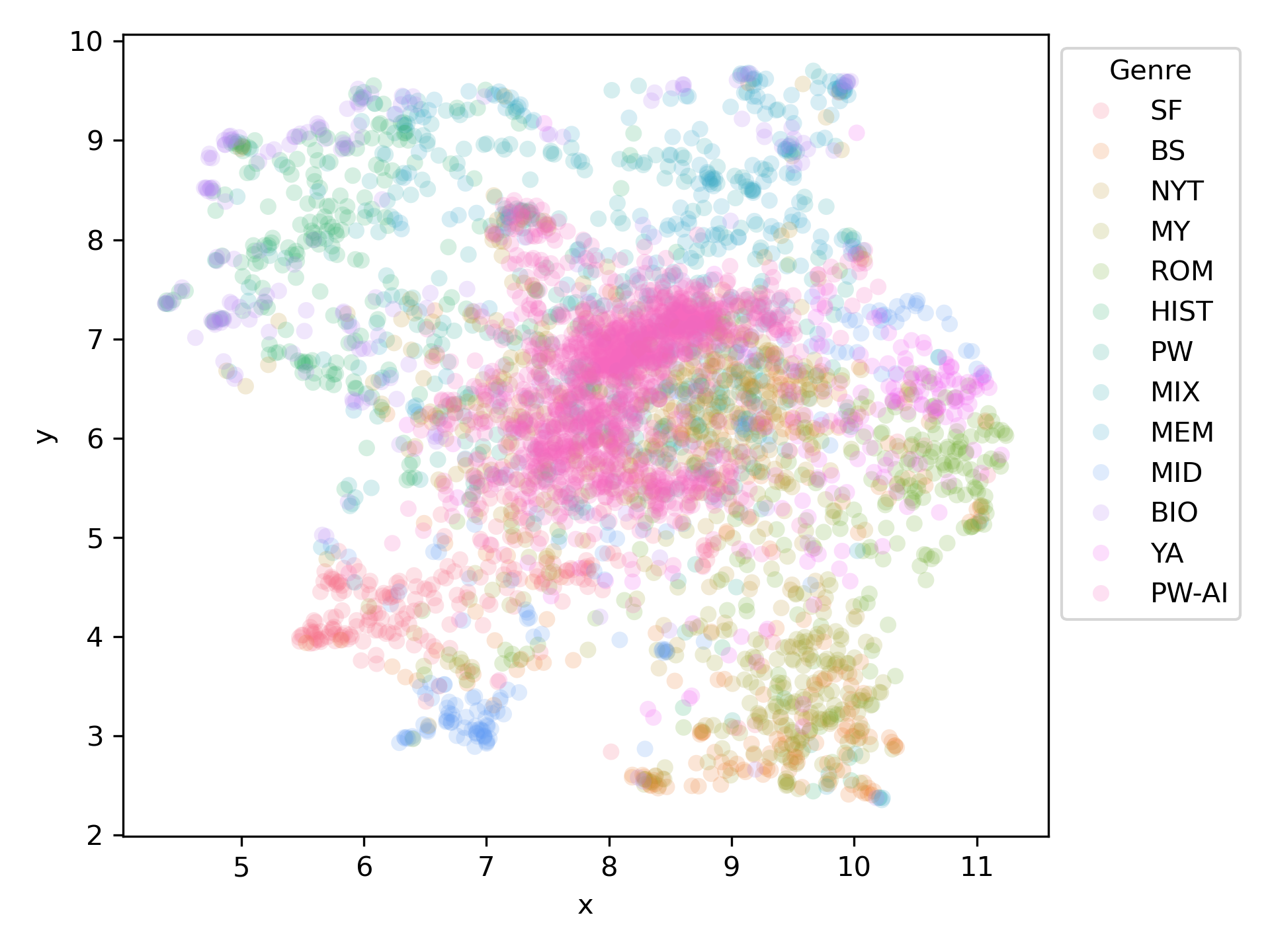}
        \caption{With AI-generated PW (prize-nominated) texts}
        \label{fig:second_sub}
    \end{subfigure}
    
    \caption{Visualization of narrative texts in dimension-reduced embedding space.}
    \label{fig:genre-map}
\end{figure*}

The first question is whether AI-generated texts preserve the genre structure observed in human fiction. Figure~\ref{fig:genre-map} visualizes the distribution of genres among human-authored texts in CONLIT, as well as the superposition of AI-generated texts targeting the prize-nominated genre. Table~\ref{tab:genre_sim} summarizes the average within-genre pairwise cosine similarity of human-authored and AI-generated texts. Table~\ref{tab:within-genre} focuses on the similarities of just those genres for which complex-prompted AI-generated texts were produced.

\begin{table}[ht]
\centering
\small
\begin{tabular}{l S[table-format=1.3] S[table-format=1.3]}
\toprule
\textbf{Genre} & {\textbf{Mean Similarity}} & {\textbf{Std Dev}} \\
\midrule
SF      & 0.653 & 0.055 \\
MY      & 0.647 & 0.050 \\
PW      & 0.635 & 0.059 \\
YA      & 0.632 & 0.048 \\
MID     & 0.629 & 0.058 \\
BS      & 0.617 & 0.061 \\
MY-AI   & 0.610 & 0.049 \\
ROM     & 0.604 & 0.045 \\
NYT     & 0.604 & 0.051 \\
SF-AI   & 0.601 & 0.051 \\
PW-AI   & 0.580 & 0.055 \\
MEM     & 0.548 & 0.077 \\
BIO     & 0.485 & 0.071 \\
HIST    & 0.479 & 0.080 \\
MIX     & 0.467 & 0.074 \\
\bottomrule
\end{tabular}
\caption{Pointwise mean and standard deviation of measured embedding cosine similarity by genre, including three AI-generated genre targets (*-AI).}
\label{tab:genre_sim}
\end{table}

\begin{table}[t]
\centering
\small
\begin{tabular}{lcc}
\toprule
Genre & Mean Similarity & Std Dev \\
\midrule
SF (human) & 0.653 & 0.055 \\
MY (human) & 0.647 & 0.050 \\
PW (human) & 0.635 & 0.059 \\
MY-AI (complex) & 0.610 & 0.049 \\
SF-AI (complex) & 0.601 & 0.051 \\
PW-AI (complex) & 0.580 & 0.055 \\
\bottomrule
\end{tabular}
\caption{Within-genre mean pairwise cosine similarities for human genres and complex-prompted AI genres. AI genres show lower internal similarity than their human counterparts.}
\label{tab:within-genre}
\end{table}

Three elements are noteworthy here. First, human texts have \emph{higher} within-genre similarity than their AI counterparts (PW: 0.635 vs.\ PW-AI: 0.580; SF: 0.653 vs.\ SF-AI: 0.601; MY: 0.647 vs.\ MY-AI: 0.610). In other words, AI-generated texts are, in general,  more diverse (less homogeneous) than their human-authored counterparts. This is surprising given the well-documented homogeneity of LLM outputs in prior work; we return to this point below. Second, among AI-generated texts in the target genres (prize-nominated, science fiction, and mystery), prize-nominated fiction shows the lowest within-group similarity (0.580), suggesting that the complex prompts with individualized biographies produced the most diverse outputs for this genre. Third, the rank ordering of the included genres by reception status is broadly preserved across human and AI texts, with science fiction and mystery texts, whether human- or AI-generated, having higher self-similarity than prize-nominated texts. We note, however, that human-authored prize-nominated texts are relatively self similar (that is, less diverse) compared to other human-authored genres, including even relatively low-status genres such as bestsellers (BS), young adult fiction (YA), and romance novels (ROM). This result is consistent with the earlier findings of \citet{wilkensGenreComputationVarieties2016}.

Cross-genre similarities reveal that AI-generated texts are most similar to their human genre counterparts. PW-AI is closest to human PW (0.541) among all human genres, well above its similarity to, for example, human mystery (0.500) or romance (0.467). Nearly all genre pairs are statistically distinguishable at $p < 0.001$. The UMAP visualizations shown in Figure~\ref{fig:genres-and-complexity} confirm substantial overlap between AI and human genres in the case of complex prompts, with much worse overlap in both location and dispersion in the case of basic prompts.

\subsection{Effects of prompt complexity}

A key experimental variable is the complexity of the generation prompt. Table~\ref{tab:prompts} compares within- and cross-genre similarities for human texts, complex-prompted AI texts, and basic-prompted AI texts (GPT-5 only).

\begin{figure*}
    \centering
    \includegraphics[width=\linewidth]{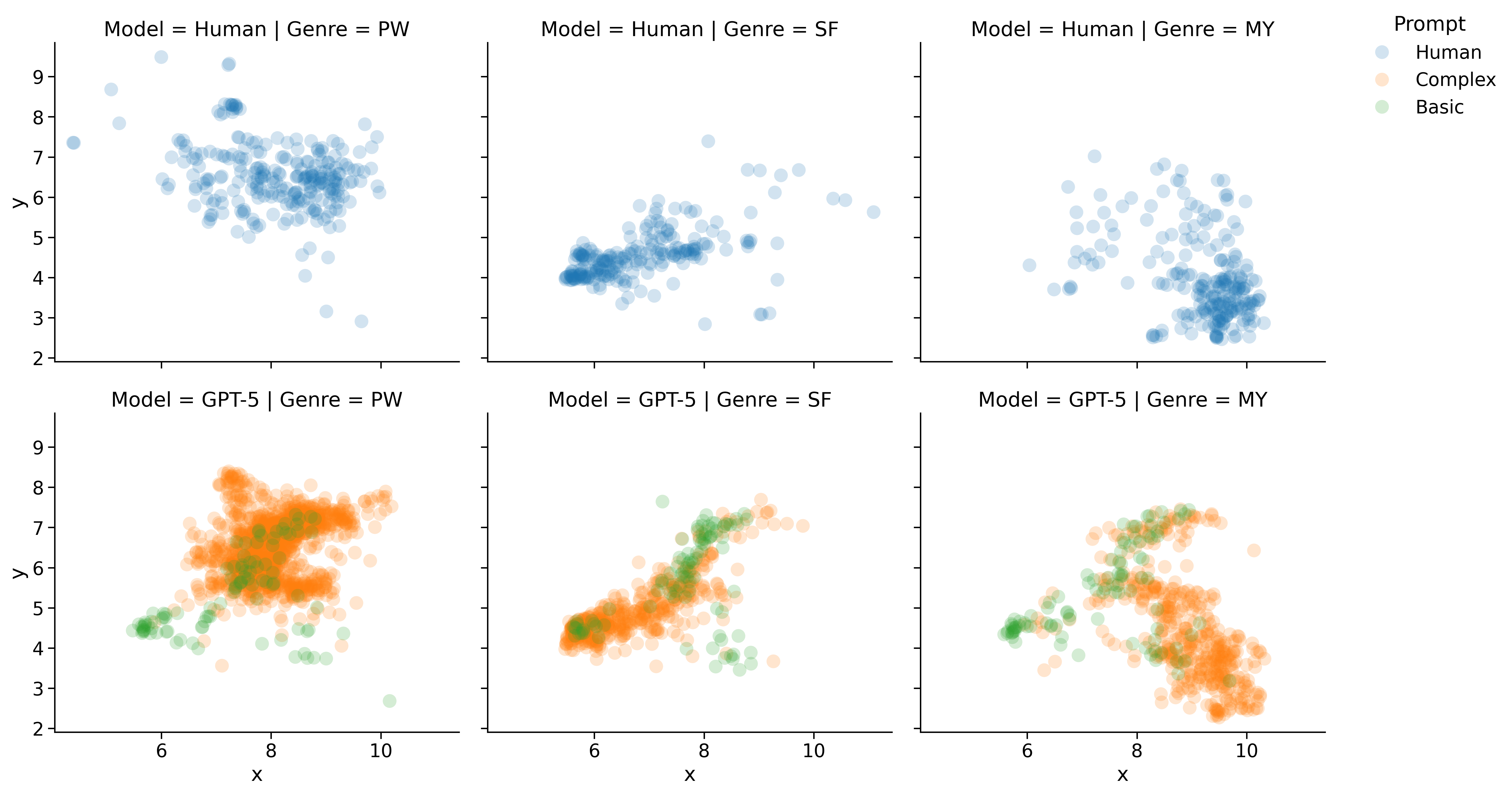}
    \caption{Comparison of human-authored texts in selected genres to AI generations under complex and basic prompting strategies.}
    \label{fig:genres-and-complexity}
\end{figure*}

\begin{table}[t]
\centering
\small
\begin{tabular}{llcc}
\toprule
Genre & Condition & Self Sim & Sim to Human \\
\midrule
PW & Human & 0.635 &  --  \\
PW & Complex & 0.580 & 0.541 \\
PW & Basic & 0.682 & 0.525 \\
\midrule
SF & Human & 0.653 &  --  \\
SF & Complex & 0.601 & 0.554 \\
SF & Basic & 0.665 & 0.511 \\
\midrule
MY & Human & 0.647 &  --  \\
MY & Complex & 0.610 & 0.564 \\
MY & Basic & 0.657 & 0.486 \\
\bottomrule
\end{tabular}
\caption{Within-genre mean pairwise cosine similarity similarity (Self Sim) and similarity to human texts (Sim to Human) for human, complex-prompted, and basic-prompted generations. All comparisons significant at $p < 0.001$.}
\label{tab:prompts}
\end{table}

The results are striking. Basic prompts produce more internally homogeneous outputs than complex prompts across all three genres (PW basic: 0.682 vs.\ PW complex: 0.580; SF: 0.665 vs.\ 0.601; MY: 0.657 vs.\ 0.610). At the same time, complex prompts are consistently closer to human texts than basic prompts (PW: 0.541 vs.\ 0.525; SF: 0.554 vs.\ 0.511; MY: 0.564 vs.\ 0.486). The gap is largest for mystery fiction, where basic prompts achieve only 0.486 similarity to human texts versus 0.564 for complex prompts.

The UMAP visualizations shown in Figure~\ref{fig:genres-and-complexity} make this pattern clear. Basic-prompted texts form  tight, compact clusters clearly distinct from the broader distributions of both human and complex-prompted AI texts. This pattern is consistent across all five models tested: GPT-3.5-turbo produces the most tightly clustered outputs, while GPT-5 with basic prompts (shown) produces slightly more dispersed outputs than smaller models, but still far more clustered than complex-prompted GPT-5 or human texts. Prompt engineering appears to affect substantially  simulation fidelity, and the design of generation prompts is a first-order methodological concern for any simulation-based research program.

\subsection{Temporal conditioning}

A subset of the complex-prompted prize-nominated chapters received a historical date directive instructing the model to write as if it were a specific year between 2005 and 2016, avoiding information after that date. The question is whether this temporal conditioning produces measurable differences in the generated texts.

It does not. The mean within-group similarity for dated texts is 0.580 and for undated texts is 0.581; the cross-similarity between the two groups is 0.580. Neither comparison approaches statistical significance ($p > 0.26$).

This null result is important. It suggests that prompt-level temporal conditioning -- simply telling the model to write as if it were a particular year -- is insufficient to produce measurably different outputs, at least within the narrow date range tested and at the resolution of the embedding space used here. This finding is consistent with \citet{underwoodCanLanguageModels2025}, which found that prompting with period prose does not produce period-consistent output, and with \citet{fittschenPretrainingLanguageModels2025}, which argued that only period-specific pretraining reliably restricts a model's outputs to a given time.

The null result also reflects, in part, the limited temporal variation in the human reference data. Comparing pre-2011 and post-2011 human prize-winning fiction reveals very little change: the average within-group similarities for the two periods differ by only 0.001 and their mean between-group similarity differs by just 0.004. If human literary fiction changes slowly over a decade, then the failure of very simple prompt-level conditioning to produce temporal differences may partly reflect the stability of the target distribution rather than solely the inadequacy of the conditioning method.

\subsection{Lexical analysis}

\begin{figure*}
    \centering
    \includegraphics[width=0.7\linewidth]{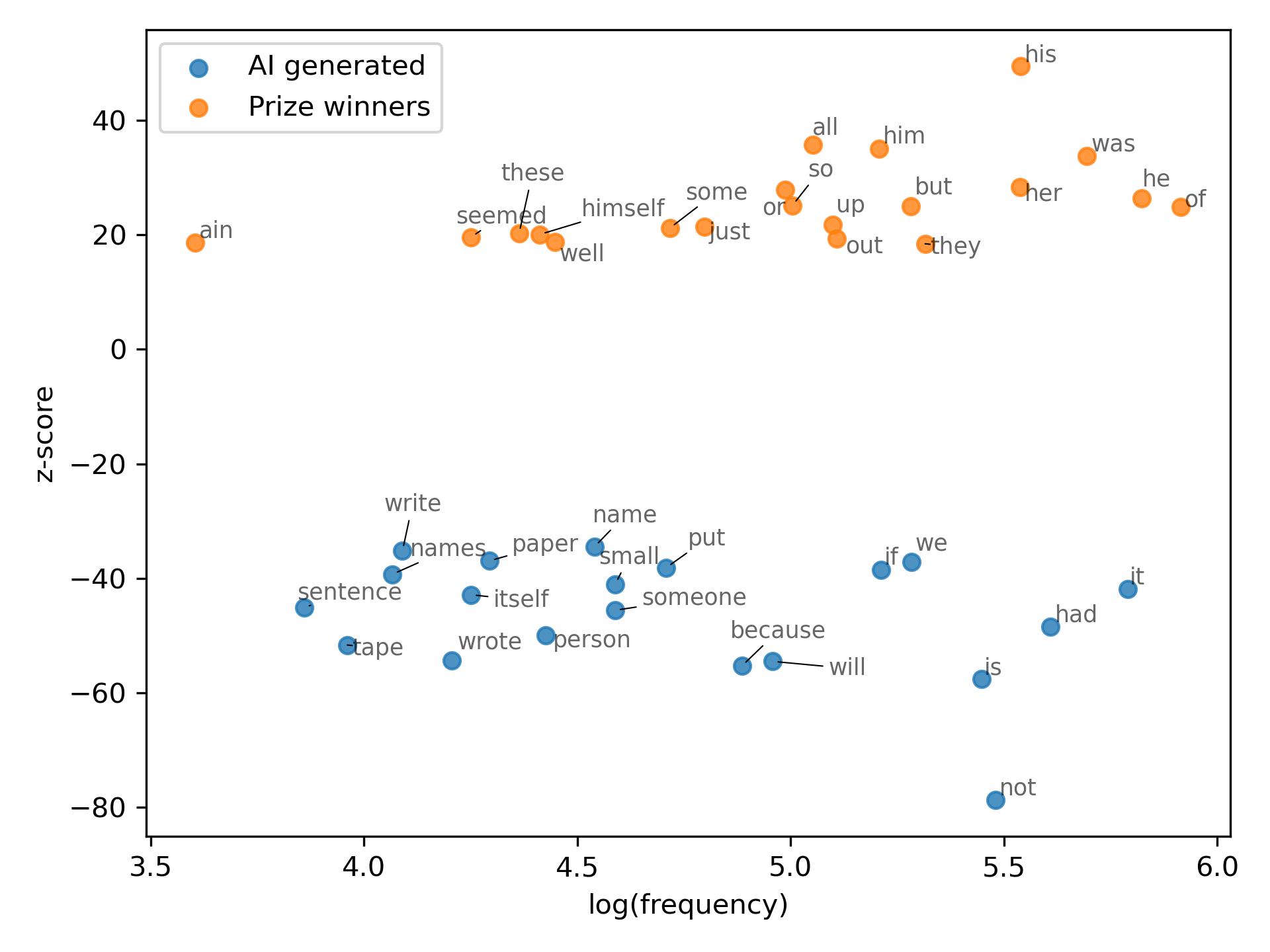}
    \caption{Fightin' Words analysis of the most distinctive words in human- and AI-authored PW fiction.}
    \label{fig:fightinwords}
\end{figure*}

To examine the character of the differences between human and AI texts at a more granular level, we applied the Fightin' Words method of \citet{monroe2008fightin} to compare 258 human prize-nominated novels against 1,400 complex-prompted PW-AI chapters, using 1,676 non-prize fiction volumes as a prior. The results of this analysis are shown in Figure~\ref{fig:fightinwords}.

The results point in two different directions. On one hand, the observed lexical differences are modest compared to those often seen for clearly separated groups (Democratic vs.\ Republican Congressional speeches, for example). In the present case, the most distinctively distributed terms are mainly pronouns and function words. If there is a clear and substantial difference in subject matter or style between human-authored and AI-generated texts, it is not immediately apparent in these results.

That said, there do exist detectable differences in register and narrative mode. Human prize-winning fiction is characterized by third-person pronouns (\emph{his}, \emph{him}, \emph{her}, \emph{he}), past-tense markers (\emph{was}, \emph{seemed}), and informal or colloquial language (\emph{just}, \emph{well}, \emph{ain't}, \emph{so}). These features point to embodied, third-person, past-tense narration with represented speech and dialogue.

AI-generated texts contain more present-tense markers (\emph{is}, \emph{will}), negation (\emph{not}, the single most distinctive AI word with $z = -78.6$), and a cluster of possibly metafictional terms that includes \emph{wrote}, \emph{sentence}, \emph{write}, and \emph{paper}. They also show more explicit logical connectives (\emph{because}, \emph{if}) and the first-person plural (\emph{we}).

Previous work has found limits to embodiment in AI-generated text \citep{hicke2025zerobodyproblemprobing}, as well as a preference for first-person plural expression \citep{walshDoesChatGPTHave2024}. There may be evidence in the present results that AI-generated literature, even when produced via a complex prompting strategy and with a state-of-the-art model, continues to exhibit some of the characteristics conventionally associated with AI literary texts.

\section{Discussion}
\label{sec:discussion}

Our experimental results establish several points relevant to the viability of simulation-based literary-historical research. AI-generated literary fiction is measurably similar to, but distinguishable from, human literary fiction in embedding space. Genre structure is preserved; AI texts generated toward prize-nominated, science fiction, and mystery targets are more similar to their human genre counterparts than to any other genre. Prompt engineering has a large effect on simulation fidelity; complex prompts with synthetic biographies produce more diverse and more human-proximate outputs than basic prompts. But gaps remain: AI texts can are lexically distinguishable from human texts in ways that reflect differences in narrative register, and minimal prompt-level temporal conditioning produces no measurable effect.

The overall result is one of partial but informative simulation. The method captures some structural features of the literary field (genre organization, the effects of authorial individualization) while missing others (aspects of register, presumed temporal specificity). This is analogous to early-stage simulation in other fields. The first climate models captured large-scale atmospheric circulation, but not regional precipitation patterns. What matters is not (yet) whether the simulation is complete, but whether it is directionally useful and improvable.

\subsection{Overcoming homogeneity}

A central concern about using LLMs for literary simulation is that their outputs are too homogeneous to represent the diversity of actual literary production. This concern is well founded; a substantial literature cited in section \ref{sec:background} documents stylistic uniformity in LLM outputs. The present results complicate this picture.

Under complex prompting with individualized biographies, AI-generated texts targeted toward prize-nominated fiction show lower within-group similarity (0.580) than AI texts under basic prompting (0.682). The reduction in homogeneity is substantial and consistent across all three genres tested. This suggests that the homogeneity problem is not solely a property of the model. Indeed, newer and more capable models are not notably superior to older models when they are used in conjunction with a basic prompting strategy. The basic prompt ``Write the first chapter of a literary novel'' invites the model to produce its default conception of what such a chapter should contain. The complex prompt, with its specific authorial persona, genre expectations, quality instructions, and planning approach, appears to provide enough individuating context to push the model away from its mode and toward the tails of its output distribution.

Complex prompting does not entirely eliminate the problem, however. Even under the more sophisticated approach used here, AI-generated texts do not perfectly overlay the embedding space of human texts belonging to the same genre. But our results suggest a research direction. If simple biographical context reduces homogeneity by over 0.1 in mean pointwise cosine similarity, more sophisticated conditioning methods -- fine-tuning, style transfer, multi-model ensembles -- may reduce it further. And we note that AI-generated texts show \emph{greater} average within-genre diversity than their human-authored counterparts. While this suggests another sense in which these texts are imperfect proxies for human-authored literature, it seems likely that it will be easier to decrease such diversity than to increase it. In any case, the question is empirical and tractable.

While we have not conducted formal ablation studies to assess the impact of each part of our generation strategy, we can measure the mean similarity of texts generated using the same synthetic biography (since we randomly sample the set of 233 biographies based on unique PW authors 1,400 times in the course of generating our complex-prompted samples, most synthetic biographies are used multiple times). We find that texts produced using the same biography have mean pointwise similarity to one another of 0.71, significantly higher than the observed similarities across the full corpus (which, recall, are around 0.6). This number is only very slightly higher in the case of synthetic biographies that a separate LLM can ``unmask'' by guessing the source author. While our PW dataset is too small to calculate an equivalent metric for human-authored texts (there are very few repeat authorial nominations in our prize corpus), the fact that controlling for author biography increases homogeneity is not surprising. It does, however, raise an interesting challenge for future historically oriented work, where synthetic biographies may not be so well developed as those produced for contemporary authors.

Different research questions also have different tolerances for homogeneity, of course. Studies of central tendencies -- what does the ``average'' prize-winning novel of a period look like? -- are relatively robust to a simulation that under- or over-represents outliers. Studies of innovation and outlier dynamics -- how do aesthetic innovators reshape a genre? -- require a simulation that captures the tails of the distribution. The former is currently more feasible than the latter.

\subsection{Counterfactual conditions and causal inference}

The paper's most ambitious claim is that simulation could eventually support causal inference based on counterfactual literary histories, including simulation-based experiments that ask how literature would have differed if some condition had been otherwise. The present results provide both encouragement and caution.

Encouragement comes from the demonstration that conditioning affects outputs in predictable ways. The difference between basic and complex prompts is not random noise; it reflects the systematic effect of providing more specific context. If a model's outputs can be steered by biographical information and genre instructions, then in principle they can be steered by other kinds of conditioning, including historical context, cultural knowledge, and the presence or absence of specific influences.

Caution derives from the dated/undated null result, which shows that very simple prompt-level temporal conditioning, at least as implemented here, is insufficient. This is consistent with \citet{underwoodCanLanguageModels2025}, which found that prompting alone does not produce period-consistent text. The implication is that genuine counterfactual simulation will require interventions at a deeper level than the prompt. Pretraining exclusively on period documents \citep{IntroducingTalkie13B} presents one option, albeit an expensive one. Model editing \citep{mengLocatingEditingFactual2023, mengMassEditingMemoryTransformer2023} offers another approach, directly altering the factual associations in a model's weights to create a version of the model that ``knows'' a different literary history. Model unlearning \citep{eldanWhosHarryPotter2023, mainiTOFUTaskFictitious2024} offers another, removing specific knowledge to create a model that behaves as if certain texts or authors were never part of its training data.

Each of these approaches faces significant technical challenges. As noted in section~\ref{sec:background}, full pretraining is expensive, faces an upper bound on available data, and does not guarantee historical localization (on this last issue, see also \citealt{breenAreVintageLLMs2026}). Current editing methods work best for discrete factual associations, not for the distributed stylistic and cultural influence that authors, movements, and institutions exert across entire literary cultures or eras. Current unlearning methods, as the TOFU benchmark demonstrates, do not yet achieve the kind of thorough knowledge removal that would be needed for a clean counterfactual. But these are challenges, not necessarily impossibilities. The combination of model editing, unlearning, and period-specific fine-tuning or pretraining defines a plausible if demanding research program for the coming years.

\subsection{Types and goals of simulation-based literary history}

There are hundreds or thousands of literary-critical and {\hskip0pt}\-/historical questions that simulation-based experiments might help to answer. At bottom, the class of potentially addressable problems is those that would benefit from reference to a grounded measure of how a definable cultural situation might have evolved (with or without designed intervention) over some period of time. Examples include:

\begin{enumerate}
    \item Assessing the influence of a writer, event, or movement by withholding from (or introducing into) a model knowledge of the target (whether through bespoke training or post-hoc ablation/insertion) and generating new texts to compare to those that actually exist. In shorthand: What might science fiction have been without Ursula Le Guin? How might American novels have evolved c.\ 2008 in the absence of the Great Recession, or following a hypothetical downturn of the same scale in the mid-1990s?
    \item Repeated ``rerunning'' of a modeled subset of a given historical juncture, with the goal of simulating the range of possible (and probable) outcomes, which can then be compared to existing historical observation. Again in shorthand, how surprising is the emergence of a writer like Virginia Woolf, or of the specific representational forms of Anglo-American modernism, given the state of Britain and of British literature (or just of Bloomsbury) in 1915?
    \item Intervening on modeled collections of interacting agents to study the specific cultural and aesthetic effects of conditions that cannot otherwise be studied in isolation. As noted at the outset, this might involve creating matched synthetic personas for writers, differing only in the imposition of the experience of migration or of economic precarity at a fixed life stage.
\end{enumerate}

Each of these experiment types is currently feasible, if not yet widely validated, in the contemporary case, although it is obviously difficult to compare simulated historical paths to observed ground truths over a period that may comprise only a few months or years after the knowledge cutoff for such models. Properly historical cases remain just out of reach absent high-performing, robustly steerable historical models. But those models are not far off; they are likely to be months, not years or decades, away.

Beyond the technical challenge of historical simulation, what are the true aims of such an undertaking? In each of these cases, a goal may certainly be to understand better the specific situation in question; many of these questions and problems are important in their own right to literary historians. But they will also contribute to broader models of the configurations, networks, developments, and events that are strongly or weakly conditioned by social context. Such models are key objects of historical reasoning. They are the area in which simulation is likely to have the greatest impact insofar as simulation offers newly grounded evidence concerning the probability distributions of historical change. In other words, a robust program of simulation will in aggregate help scholars reason about the causal mechanisms of history as much as (or more than) any simulation experiment in isolation will illuminate its direct object.

\subsection{Limitations and future work}

\subsubsection{Cultural specificity and historical fidelity}

The experiments reported here focus on contemporary, English-language literary fiction, precisely the domain where LLM training data are richest and where simulation is therefore most likely to succeed. Indeed, this is why we selected the domain and dataset we did in pursuit of a proof of concept. Any extension of the method to other literary traditions, historical periods, or languages will encounter additional challenges.

Instruction-tuned LLMs are further aligned to the values of similar markets by predominately similarly situated corporations and AI labs. Their cultural knowledge of past and non-Western literary traditions may be thin, stereotyped, or absent \citep{chiuCulturalBenchRobustDiverse2024a}. \citet{veselovskyLocalizedCulturalKnowledge2025} show that, while local cultural information persists within LLMs, it does not surface naturally, and steering toward specific cultural contexts risks stereotyping (see also \citealt{lee2026pooralignmentsteerabilitylarge}). \citet{liuItBadWork2025} find that GPT-4 generates social norms that are less culture-specific than those of human populations, a result supported by \citet{hobsonStoryMoralsSurfacing2024} and \citet{wu2026lessonsbordersevaluatingcultural}. These findings suggest that cultural flattening is a risk for any simulation that attempts to model literary production outside the well-represented Anglo-American mainstream. They may also suggest that base models will be preferred to instruction-tuned versions in some contexts, despite the more limited instruction-following capabilities of the former.

The anachronism problem remains a concern. If a model cannot reliably produce texts that resemble fiction from a specific historical period, then simulating what that period's literature might have looked like under different conditions is doubly problematic, since the baseline itself is unreliable and any counterfactual comparison inherits that unreliability. Work is underway to develop benchmarks of historical fidelity not just in style but in reasoning \citep{Underwood_Qiu_Griebel_Nelson_Roland_Shang_Wilkens_2026}. These evaluation metrics will be a key part of any future literary-historical simulation research. As noted above, there remains an important if necessarily partial role for human expert evaluation in the benchmarking process.

The above points are genuine limitations of power and scope. The method is currently most viable for contemporary English-language fiction well-represented in training data. It becomes progressively less reliable as one moves toward historical, non-Western, or minority-language traditions. This does not invalidate the approach, but it constrains its applicability in ways that researchers must take seriously. Period-specific pretraining, despite its difficulty (largely due to the relative scarcity of training data in even the medium-distance past) and cost, may be a prerequisite for some historical simulations.

\subsubsection{Other limitations}

Several technical limitations should be noted. Generated texts used in this work are single first chapters of approximately 10,000 words, not full novels. While initial-chunk embeddings are validated here as stand-ins for full-book embeddings, it remains the case that we compare parts of (nonexistent) AI-generated novels to complete human-authored novels. Future experiments should investigate full-length novel generations, though these will be significantly more expensive to produce and to validate (especially in cases that require human reading). 

All complex-prompted texts use a single generative model (GPT-5). These results may not generalize to other frontier models. The current work does not seek to compare the performance of different models on the target task, merely to demonstrate that at least one model and approach is able to produce plausibly in-distribution AI-generated texts resembling selected genres of human literary writing. An obvious next step is to repeat this process with other models and to assess the differences in their performance.

The embedding-based analysis captures style, topic, and broad narrative form, but does not assess directly narrative quality, coherence, sustained character development, or the kind of interpretive richness that many literary scholars value. Two avenues suggest themselves in this case. First, there exist dozens of NLP systems that seek to measure aspects of narrative (for overviews, see \citealt{hamilton-etal-2026-narrabench} and \citealt{russellStoryScopeInvestigatingIdiosyncrasies2026}). These might be applied to both the human-authored and AI-generated texts used in this research to understand more fully the areas in which the two sets of texts do and do not converge. Second, methods of mechanistic interpretability might be applied directly to the embedding space so as to map the differences between specific regions of that space.

In a related vein, the Fightin' Words analysis, while informative, operates at the token unigram level and does not capture syntactic patterns, narrative structure, or semantic coherence. A more complete validation framework, perhaps using an LLM-as-judge approach \citep{zheng2023judging}, is likely to offer more robust evidence concerning the qualitative differences between the two sets of texts.

We presume that there remains significant room for improvement in the overall quality of the generated texts. In part, this is simply a matter of trying more models, prompting strategies, and other details of the current approach. This process might itself be automated by using LLMs to suggest and evaluate new prompts that narrow the gaps between human and AI texts.

Finally, the present results do not attempt to perform full-blown simulation-based causal inference, even setting aside historical conditioning. This is a clear direction for future work.

\section{Conclusion}
\label{sec:conclusion}

This paper has described a methodological framework for simulation-based experiments in literary studies, grounded in the analogy between AI-generated text and the computational simulations used in the natural and social sciences. The framework has been validated through a large-scale comparison of 3,800 AI-generated fiction chapters against a corpus of human-authored literary novels, using embedding similarity and lexical analysis. It presents the first evidence of which we are aware that it is possible to generate, at relatively low cost, in-distribution synthetic samples of human-authored literature, including in genres defined by high reception status.

The results establish that AI-generated literary fiction preserves meaningful genre structure, that prompt engineering substantially affects simulation fidelity, and that individuated authorial context (via synthetic biographies) reduces the homogeneity that characterizes default LLM outputs. They also reveal significant remaining limitations including systematic differences in narrative register between AI and human texts, the failure of simple prompt-level temporal conditioning, and a persistent homogeneity gap (albeit in the opposite and more tractable direction than that found in prior work). We understand these limitations as findings that define the method's current capabilities rather than evidence of fundamental impossibility.

The most important technical advances needed for this research program are three: better counterfactual conditioning (through model editing, unlearning, or period-specific pretraining), further improvement in output steerability and reduction in homogeneity (through more sophisticated prompt design, fine-tuning, or multi-model approaches), and deeper validation metrics in addition to embedding-level similarity to assess narrative structure and qualitative features.

This paper is intended as an invitation. For literary scholars, it offers a new class of research tools -- imperfect but improvable -- for addressing the counterfactual questions that have always animated literary history. For NLP researchers, it identifies a challenging application domain that requires advances in model editing, cultural conditioning, and output diversity. The method is already tenable for limited use cases. We believe it is on the edge of usability for more sophisticated cultural and historical simulations. We look forward to collaborating with researchers across disciplines to develop further robust, validated, and adaptable systems of AI-based experimental literary simulation.

\bibliography{bibliography}

\appendix
\section*{Appendix}
\label{sec:appendix}
\section{Prompt examples}
\label{sec:prompts}

The examples below are tuned to the prize-nominated (PW) genre case. Each is modified as appropriate for other genre targets. Instances in which variables are supplied in code are indicated with \{curly braces\}. The full text of all prompts, as well as the generated biographical sketches and output chapters, are available in the \href{https://github.com/wilkens/literary-generation}{code repository} for this article.

\subsection{System prompt}
\begin{quote}
    You are an expert literary novelist. Your objective is to craft stories that are emotionally and intellectually engaging, formally innovative, and thematically original. Your stories should appeal to literary critics and discerning readers. Begin your reasoning with a concise checklist (3-7 bullets) outlining your creative approach for each new story; keep items conceptual rather than implementation-specific. Do not include the content of the checklist in your output. You possess a strong familiarity with recent winners of major literary prizes, drawing inspiration and insights from those works without direct imitation. Maintain your own distinctive narrative voice and style at all times, never mimicking an existing author. Set reasoning\_effort = medium to balance creativity and depth without unnecessary verbosity.
\end{quote}

\subsection{User prompt}
\begin{quote}
    Personal context: \{bio\}. Historical context: Respond as if you are writing in the year \{year\}, avoiding information or events from later years. The task: Write the first chapter of a new novel that might be competitive for a literary prize. The chapter should be about \{wordcount\} words long. It should not rely on clichés, neither in form, nor in style, nor in content. It should be devoted to a theme that is important to you and that is typical of prizewinning literary fiction, though it may also cover new ground. It may use any form and style that supports its expressive goals. Begin by making a careful plan or outline of the novel as a whole, as well as a description of its principal characters. Use this plan and character sheet to inform the content of the chapter you are writing. Do not count the plan as part of the chapter; that is, write the full, \{wordcount\}-word chapter after you complete the plan. Do not output the content of the plan or the character sheet. Return only the content of the chapter.
\end{quote}

\subsection{Synthetic biography generation prompt}
\begin{quote}
    Write a short biographical summary of an imaginary author who closely resembles the novelist \{author\}. Your biography should be about 100 words long. It should be written as if the current year were 2010, excluding information from later years. It should be written in the second person, as if addressed directly to the imaginary author. It should maintain a serious tone, avoiding slang and clichés. The biography should state the imaginary author's year and place of birth, summarize their life and career, describe the distinguishing characteristics of their work, and list at most one or two major awards they have won. These details should NOT be exactly the same as those of \{author\}, but they should be similar and they should be realistic. Do not invent fictitious places, awards, or book titles. Do not give the imaginary author a name. Do not mention the current year.
\end{quote}

\subsection{Embedding prompt}
\begin{quote}
    You are an expert literary critic. Consider the following excerpt from a prize-winning novel published in the twenty-first century. Create a comprehensive embedding that incorporates the text's subject matter, style, and narrative form. The embedding will be used for clustering.
\end{quote}

\end{document}